# Reward Engineering for Object Pick and Place Training

Raghav Nagpal[1], Achyuthan Unni Krishnan[2] and Hanshen Yu[1] .

*Abstract*—Robotic grasping is a crucial area of research as it can result in the acceleration of the automation of several Industries utilizing robots ranging from manufacturing to healthcare. Reinforcement learning is the field of study where an agent learns a policy to execute an action by exploring and exploiting rewards from an environment. Reinforcement learning can thus be used by the agent to learn how to execute a certain task, in our case grasping an object. We have used the Pick and Place environment provided by OpenAI's Gym to engineer rewards. Hindsight Experience Replay (HER) has shown promising results with problems having a sparse reward. In the default configuration of the OpenAI baseline and environment the reward function is calculated using the distance between the target location and the robot end-effector. By weighting the cost based on the distance of the end-effector from the goal in the x,y and z-axes we were able to almost halve the learning time compared to the baselines provided by OpenAI, an intuitive strategy that further reduced learning time. In this project, we were also able to introduce certain user desired trajectories in the learnt policies (city-block / Manhattan trajectories). This helps us understand that by engineering the rewards we can tune the agent to learn policies in a certain way even if it might not be the most optimal but is the desired manner.

## I. INTRODUCTION

### A. Significance of Reinforcement Learning in Robot Reach-to-Grasp

Robot reach-to-grasp is one of the top applications of robot manipulators[9], especially in industrial use such as warehouse management and automobile manufacturing[4]. According to Reuters and Robotic Industries Association's data, 35580 new robots are shipped to the US and Canada, in all industries just in the year 2018. Among them, over 60% are used for handling and grasping on the assembly line, making reach and grasp the most important component of all industrial robot applications.

The theory of grasping manipulation starts from Reuleux's book, the kinematics of machinery, written in 1875[24][16]. In late 1970s C-space approach was raised by Lozano-Perez, Mason and Taylor, which made up the basis of robot motion planning to solve the problem of how to reach and grasp[11]. The early works made autonomous robot grasping possible. In late 1980s, Roth and his team tried and solved the basic problems in robot grasping and dexterous manipulation, base on the classical line geometry approach[17].

Two methods of robot reach-to-grasp motion planning can be categorized as, classical and heuristic.[14]. The classical methods are stable, however they require a time intensive computations and would be trapped in the local minima[12]. Thus we need the heuristic approaches to emphasize exploitation and reduce computational complexity[18]. One of the efficient algorithms is nonlinear model predictive control, addressed by Hsieh[7], focusing on dynamic known environments.

Reinforcement learning techniques are then used in solving this problem[8]. By incorporating reinforcement learning algorithms, robots could learn reach-to-grasp on a new or dynamic environment, thus creating a need on more strong and efficient algorithms, their alternates or simply some improvements on learning a model.

### B. Motivation for the Project

Since it is a great challenge for robotic grasping under uninstructed environments[10], most of the robot motion is now planned manually for each individual task before deployment for labour. Despite the concerns of safety and failures, the offline training of a reinforcement model is also costly in both computational effort and time. It is also extremely limited with how robust the agent can be in terms of executing tasks. One way to make the process quicker is to shape better reward functions than the normal heuristic ones. Thus our project aims to compare the reach-to-grasp training performance, based on multiple reward functions.

| Objective | Description |
|---|---|
| Paper review | Learn about prior state-of-the-art and possible methods. |
| Train a baseline model | Get familiar with OpenAI simulation environment(used traditional hard coded heuristic based motion planning), finish the code and train a baseline model. |
| Execute multiple reward functions | Compare the training performance of different models: success rate and convergence time. |
| Generate trajectories from the trained models | Compare the trajectories based on different reward functions. |
| Finish report and present work | Complete the report, poster and other paper work |

TABLE I: Project objectives.

## II. RELATED WORK

### A. Deep Reinforcement Learning

Motion planning with distance as a simple heuristic was used to move the robot to the object and then the goal. The

[1] Raghav Nagpal and Hanshen Yu are with the Robotics Engineering Program, Worcester Polytechnic Institute, Worcester, MA 01609, USA {raghav,hyu5}@wpi.edu
[2] Achyuthan Unni Krishnan is with the Mechanical Engineering Department, Worcester Polytechnic Institute, Worcester, MA 01609, USA aunnikrishnan@wpi.edu

Please find the link for the video demonstration here:https://www.youtube.com/watch?v=CFepYCdxlQs . The Github repository for this project can be found here: https://github.com/raghavnagpal/baselines

planner failed when picking up the object when approaching from certain angles. Reinforcement Learning solves this problem by innately learning the strategies.

This problem is a continuous action state-space. This implies that there are infinite possible configurations or a lot of state spaces to explore. This means conventional Reinforcement algorithms fail as it is impossible to cover such a large set of possibilities.

*B. Reward Shaping*

Reward shaping is a technique to make a model easier to learn, by adding or modifying the reward function in reinforcement learning. Usually, there are obvious heuristics, such as the target in reach-to-grasp problems, and a win for games[20]. Ng proposed a method for adding shaping rewards in a way that guarantees the optimal policy maintains its optimality[15]. Based on this multiple reward shaping methods are proposed such as potential based[21] and plan based[6], for different settings[3], and proved by theoretical analyses[5][13].

For this reach-to-grasp problem, by using a shaped reward to bias the learned model, the robot could reach to the goal more efficiently. We know an intuitive way of going to a goal is reduce the distance to the final goal. The best way to reduce reach the goal is by reducing the distances to the goal along the three dimensions so that the target goal is reached. As a result we apply the knowledge that we possess on how to solve this problem to augment reward functions and learn to reach the goal faster than otherwise.

*C. Sparse Reward Problem*

Sparse rewards, along with human demonstrations, are practical alternatives to teach robots to solve control tasks, including reach-to-grasp[19]. Sparse rewards have the problem of several experience episodes have no meaning at all as the rewards might be 0. This is where Hindsight Experience Replay (HER) plays a key role. HER lets the agent know that any wrong action can be the right action if the goal had been in the new state as a result of the wrong action[1].

Fig. 1: Pseudo-code of HER implementation.[1]

The current way in which the environment is designed is that the reward function is sparse. Meaning no reward for every moment when the goal is not achieved and only a positive reward when the goal is achieved. HER was implemented which significantly boosted the successes of the algorithm. A study of the impact of HER and it's results were published in their blog about the gym for robotics. As clearly indicated in Fig 2 DDPG with HER was the most successful implementation by a substantially considerable margin. That is the reason why we decided to utilize the baselines provided with this infrastructure to train our models with our innovations in reward functions.

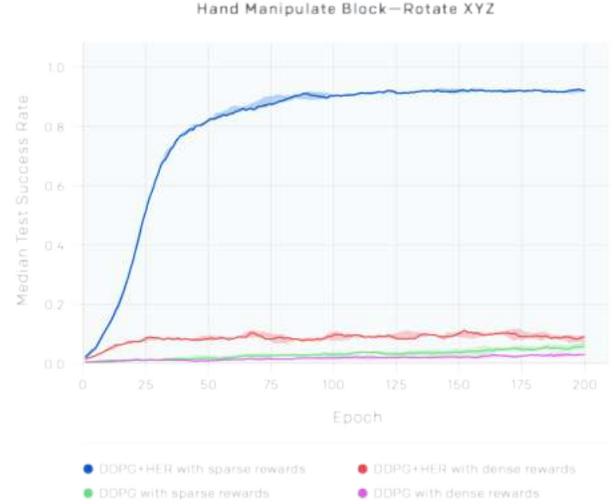

Fig. 2: Impact of HER implementation on Shadow Hand manipulation.[1]

Considering that we have prior knowledge on reach-to-grasp problems, we could take this sparse reward problem and control the policy in a certain direction. Hence in this project we seek to use the successes of HER in sparse reward situations and the DDPG model which is successful in continuous state spaces to implement a solution for the pick-and-place problem for our attempts at learning how to engineer the rewards functions.

### III. OpenAI gym environment and Baselines

For this project we take advantage of OpenAI environments and baselines[2], particularly the Robotics environment 'FetchPickAndPlace-v1' for this project. We design our rewards in the configuration file that details the environment observation state space.

*A. Environment*

The 'FetchPickAndPlace-v1' environment provides the infrastructure to train the agent to pick an object and move it to a pre-determined goal.

The agent being controlled is the arm of the Fetch Mobile Manipulator[23]. It is a 7 degree-of-freedom arm with a gripper at the end of the arm.

The environment works based on the MuJoCo physics simulator and is included in the OpenAI gym library. The following information can be extracted from the environment:
- The system state including gripper position, object position and orientation, goal position, gripper state (open or close), gripper velocity and object velocity.
- The reward for the system: usually negative cost of living and a flat +1 reward for when the goal is achieved.
- The end-effector can be moved by supplying the target location at each frame, meaning the (x,y,z) coordinates to be moved to based on current locations of the black box, gripper and goal.

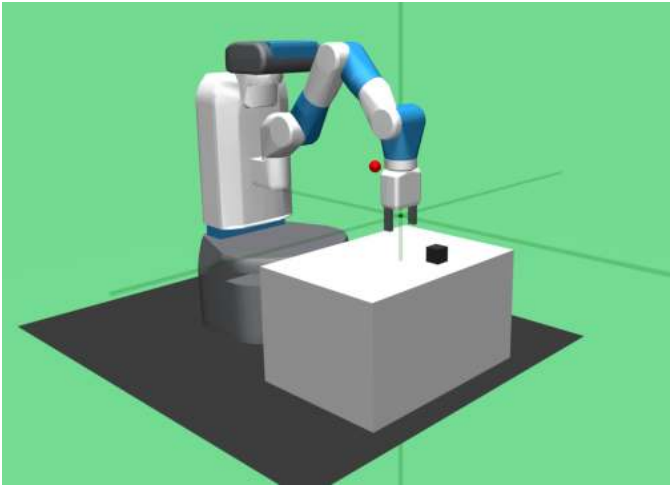

Fig. 3: An example of an episode in the environment. Red ball is the goal and the black box is the object

## B. Baselines

OpenAI provides baselines which are a set of high quality implementations of Reinforcement learning algorithms. These implementations and the performances achieved are almost on par with the results from published papers.

For this purpose the baselines were built off of to be more efficient with the progress of the project as the ultimate goal was to identify the nuances of Reward Engineering.

Training of the models were done on the following two systems. MuJoCo-py and gym were required to run the environment while the model was trained using models written in Tensorflow which takes advantage of the Nvidia parallel computing architecture Cuda 10.1.

All the models were trained for a 1000 epochs as per the recommendation from OpenAI. The model trained thus from the vanilla reward function will be compared to in terms of performance, training time and also learned trajectories.

| Specification | Computer 1 | Computer 2 |
| --- | --- | --- |
| CPU | 8th Gen Intel Core i7-8750H processor | 7th Gen Intel Core i7 processor |
| RAM | 16 GB | 16 GB |
| GPU | NVIDIA GeForce GTX 1060 Max-Q | NA |
| Video Memory | 6 GB | NA |
| OS | Ubuntu 18.04 | Ubuntu 18.04 |
| Additional Software | Tensorflow, Cuda 10.1, MuJoCo Py and OpenAI gym | Tensorflow, Cuda 10.1, MuJoCo Py and OpenAI gym |

TABLE II: System Specifications and Software Dependencies.

The reward function of the default baseline is based on the distance from the goal. Hence there is a living cost involved in the episode continuing. As a result, the agent will try to reduce this cost and this is the least only when the goal is achieved which is the only time a positive reward achieved. Hence to maximize the reward the robot has to learn to reach the goal. Improvising on this fact modifying the reward function will result in different types of trajectories or training performances can be achieved.

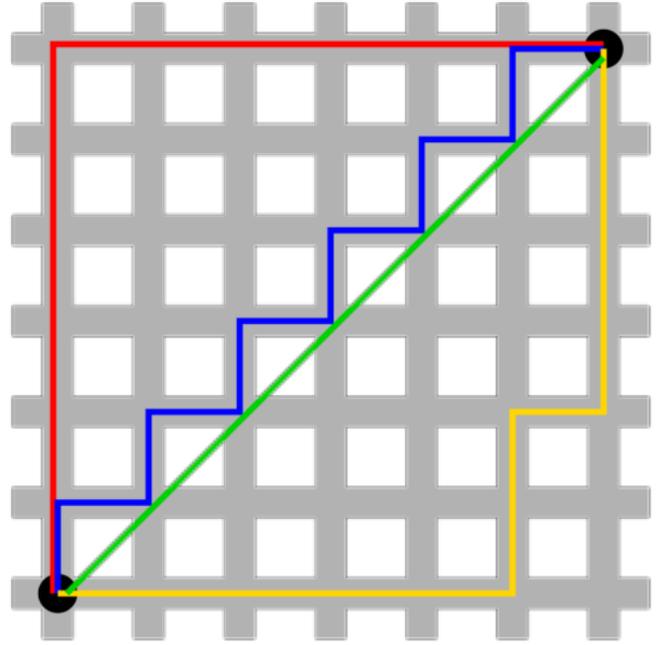

Fig. 4: The red, blue and yellow lines are examples of a Manhattan trajectory. The green line is Euclidian trajectory for comparison.[22]

## IV. REWARD ENGINEERING

The reward type is sparse, meaning the only positive reward occurs when the goal is achieved. This means in order to maximize the reward it should always try to move the black box to the red goal. Hindsight Experience Replay (HER) was used to immediately compensate for this problem of sparse rewards.

As mentioned the goal of this project is to study the process of optimal reward engineering. We also tried to induce certain characteristics into the final learned trajectories using careful designed reward functions. In this project we were able to successfully implement the final learned trajectory to be a Manhattan/Taxi-cab trajectory[22].

In Table III some of the successful reward functions we were able to implement are mentioned.

As previously mentioned the Vanilla reward function only has a negative cost of living and a final positive reward. This reward function was used to train the model that we would ultimately use as the benchmark to compare the performance, ease of training, etc of the models we trained with our modified reward functions. The reasoning behind designing our rewards as mentioned in Table III is listed below:

- *Prioritized distance from goal's z-coordinates:* The default reward function does not give any particular conditions to the agent besides reaching the goal as soon as possible. This resulted in instances where if the object or the goal was on the surface of the table the gripper tended to press down on the workspace, which is obviously not ideal. We attempted to negate this issue by increasing the penalty in the z-direction which points away from the workspace.

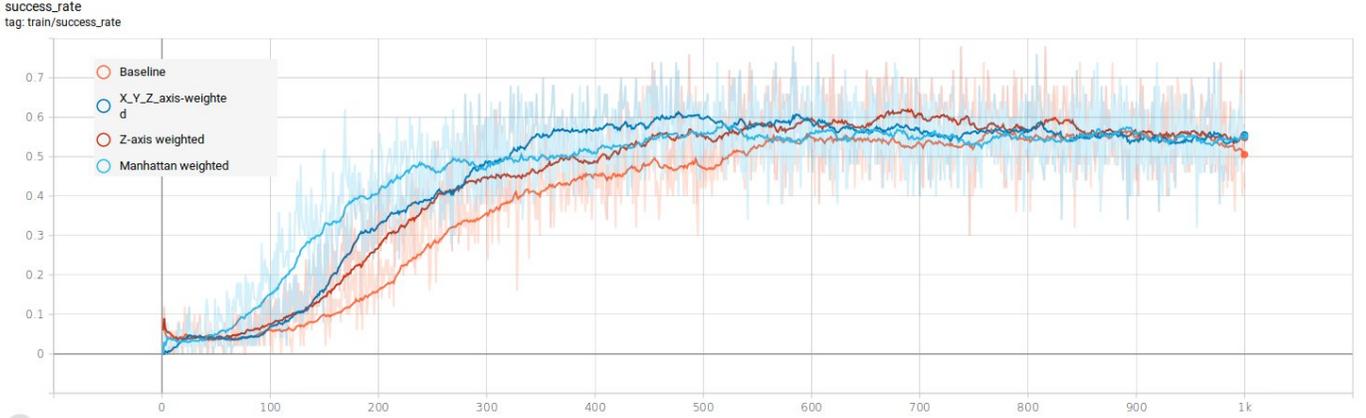

Fig. 5: Training Performances of different models. Training success vs epochs

| Reward Function | Description |
|---|---|
| Vanilla Reward Function (Model 1) | The default reward function provided by OpenAI baseline that is sparse and gives a negative cost for living with a ultimate positive reward if goal is satisfied. |
| Prioritized distance from goal z-coordinates (Model 2) | The reward function is further augmented by adding a further penalty based on the end-effector distance from goal coordinates. Higher priority was placed on the goal z-coordinate. |
| Prioritized distance from all goal coordinates (Model 3) | The reward function is further augmented by adding a further penalty based on the end-effector distance form the x,y and z-coordinates. Varying priority was placed on the goal coordinates in the following format: x-coordinate (10x), y-coordinate (5x) and z-coordinate. |
| Manhattan Trajectory (Model 4) | A constant penalty for being away from x,y and z coordinates of the goal is applied, augmenting the current reward function. Similar varied priority in the magnitude for x,y and z-coordinates can be applied. In our implementation the constant penalties were 5 in x-direction, 2.5 in y-direction and 1 in z-direction. |

TABLE III: Reward Functions.

- *Prioritized distance from all goal coordinates:* The reward function is a means of relaying what the agent is supposed to do by providing it positive and negative rewards. The way this function is designed is crucial to how effectively and efficiently the agent trains to perform the delegated objectives. In this function we decided to augment the reward function by giving an intuition of how the goal can be achieved, that is, by reducing the distance of the end effector from the goal's x,y and z-coordinates.
- *Manhattan Trajectory:* The goal of this reward is to see if we could design a reward function that could teach the agent to perform the goal exactly how we wanted it to perform the task, even if it means it is not the best solution. For this purpose we decided the agent shall complete the task in a Manhattan trajectory as previously mentioned. We gave constant penalties in the x,y and z directions if the end-effector was away from this goal. In this way, this reward function is different from the previous function as this more replicates a binary step function instead of a linear function based on distance. This means if the end effector is not at the x-coordinate of the goal it receives a constant penalty of -10 for example instead of the negative reward based on the distance. Based on the which coordinate of the end-effector needs to be equal to the goals first the weights in the reward function as shown in Table III.

## V. OBSERVATIONS AND RESULTS

### A. Training Performance

All the final models were trained for a 1000 epochs to ensure they would reach convergence. However due to different reward functions, convergence for the different models were arrived at different epochs. The training performance over time for the different models is provided in Fig 5. The model with distance from all coordinates prioritized performed the best as it reached convergence during training the fastest, while learning the Manhattan trajectory was the fastest initially but was the slowest. The default baseline provided by OpenAI was placed third. The approximate epochs when these models reached convergence is listed in Table IV.

| Reward Function | Time of Convergence |
|---|---|
| Vanilla Baseline | $\sim 650$ |
| Prioritized Distance - Only z-axis | $\sim 500$ |
| Prioritized Distance - all axes | $\sim 400$ |
| Manhattan Trajectory | $\sim 700$ |

TABLE IV: Convergence times for different reward functions.

### B. Observed Trajectories

We are also able to generate different trajectories for the same task based on the reward functions implemented. The most drastic deviation in learned trajectories is identified by using the reward function labelled Manhattan trajectory. The different trajectories developed are represented in the 3-dimensional space in Fig 6.

As portrayed in Fig 6 the baseline model, Model 3 and Model 4 all go approximately to the same coordinates. However Model 2 alone slightly deviates from this position. It is because it was trained to avoid the table workspace and as a result moves away slightly at the goal position to avoid collision.

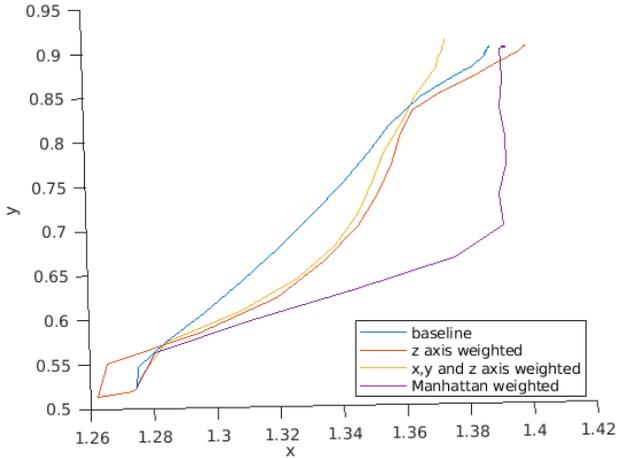

Fig. 6: Generated trajectories in 3-D space. z-axis direction is into the page.

We also observe the success of the formation of the Manhattan trajectory very clearly in the same Figure. It first seeks to achieve the x-coordinate (as this direction is weighted heavily as the initial priority) before moving to the y-coordinate and finally satisfying the z-coordinate. In order to better illustrate this feature we have split the timings at which the different x,y and z-coordinates of the goal location has been achieved in the different trajectories in Fig 7.

We observe from Fig 7 that the Manhattan trajectory results in the x-coordinate being equalled to the goal coordinate the earliest by a considerable margin, around 12 frames compared to 20 frames of the other trajectories. Additionally, it should be noted that significant motion in the z-direction is only observed after x-coordinate of the end-effector has reached a stable value. The end-effector exhibits a similar behaviour where motion in the y-direction is seen only after z-coordinate has reached a stable value.

In contrast in all the other models x,y and z-coordinates all vary simultaneously suggesting a more direct approach to the goal. This break down of the trajectories of the individual models help clearly understand that we were successful in enabling the algorithm to learn trajectories that follow a particular pattern even if it is not essentially the most optimal solution.

However not all these models were success on the first try. For the Model 3 in Table III we initially had reduced punishments for deviation from the axes. The best results were achieved when the punishments were increased by a factor of 10. On the contrary, in order to induce Manhattan trajectory these punishments had to be reduced by a factor of 10. Initially, the cost was so high the robot preferred to not move in the z-direction as the cost of deviating from the other axes were too high and as a result, limited it's exploration capabilities. These findings helped us further understand the do's and don'ts of Reward shaping.

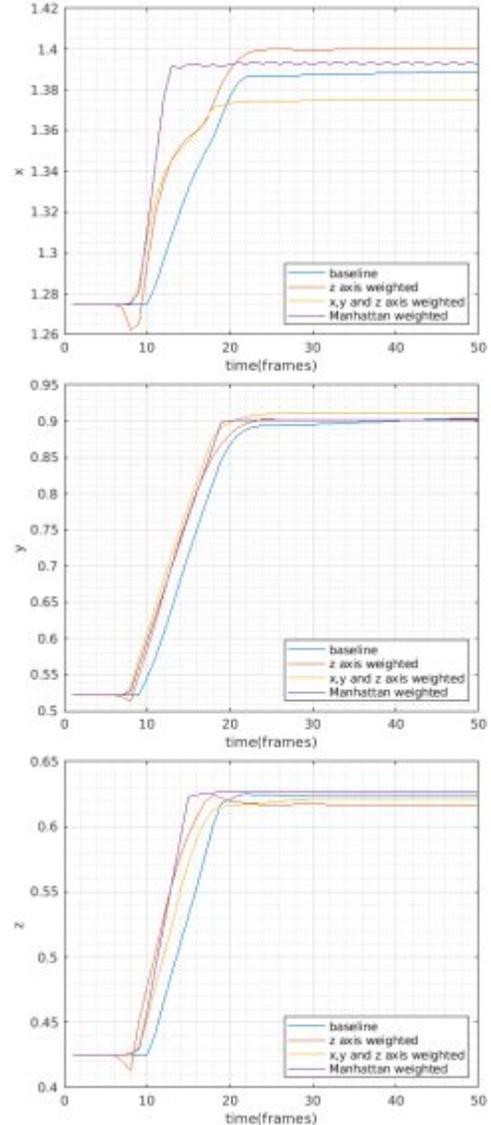

Fig. 7: First plot shows the time at which x-coordinate of goal is reached. Subsequent plots show when y and z-coordinates of the goal were reached.

## VI. FUTURE WORK

For future work, we would like to work more with the observation space and the definition of the environment itself. The current state space is only the end-effector location. Working with joint-angles may result in improved results. It can also help induce human like motion like picking up objects from the side, etc. However, in order to work within this joint space requires us to completely rework the definition of the default observation space. These modifications proved to be difficult to be incorporated within the scope of the project.

We would also like to improve training times and identify different strategies that might improve the efficiency of training. We would also like to extend these reward functions to algorithms like PPO and A2C to study their performances and if they carry over their successes to these different algorithms.

We are also intrigued in further exploring the philosophy of reward engineering and how improvements can be made to other tasks like manipulation with the Shadow Hand.

REFERENCES

[1] Marcin Andrychowicz, Filip Wolski, Alex Ray, Jonas Schneider, Rachel Fong, Peter Welinder, Bob McGrew, Josh Tobin, OpenAI Pieter Abbeel, and Wojciech Zaremba. Hindsight experience replay. In *Advances in Neural Information Processing Systems*, pages 5048–5058, 2017.

[2] Prafulla Dhariwal, Christopher Hesse, Oleg Klimov, Alex Nichol, Matthias Plappert, Alec Radford, John Schulman, Szymon Sidor, Yuhuai Wu, and Peter Zhokhov. Openai baselines. https://github.com/openai/baselines, 2017.

[3] Yang Gao and Francesca Toni. Potential based reward shaping for hierarchical reinforcement learning. In *Twenty-Fourth International Joint Conference on Artificial Intelligence*, 2015.

[4] Alessandro Gasparetto, Paolo Boscariol, Albano Lanzutti, and Renato Vidoni. Path planning and trajectory planning algorithms: A general overview. In *Motion and operation planning of robotic systems*, pages 3–27. Springer, 2015.

[5] Marek Grześ. Reward shaping in episodic reinforcement learning. In *Proceedings of the 16th Conference on Autonomous Agents and MultiAgent Systems*, pages 565–573. International Foundation for Autonomous Agents and Multiagent Systems, 2017.

[6] Marek Grzes and Daniel Kudenko. Plan-based reward shaping for reinforcement learning. In *2008 4th International IEEE Conference Intelligent Systems*, volume 2, pages 10–22. IEEE, 2008.

[7] Chung-Han Hsieh and Jing-Sin Liu. Nonlinear model predictive control for wheeled mobile robot in dynamic environment. In *2012 IEEE/ASME International Conference on Advanced Intelligent Mechatronics (AIM)*, pages 363–368. IEEE, 2012.

[8] Mohammad Abdel Kareem Jaradat, Mohammad Al-Rousan, and Lara Quadan. Reinforcement based mobile robot navigation in dynamic environment. *Robotics and Computer-Integrated Manufacturing*, 27(1):135–149, 2011.

[9] F Kamil, S Tang, W Khaksar, N Zulkifli, and SA Ahmad. A review on motion planning and obstacle avoidance approaches in dynamic environments. *Advances in Robotics & Automation*, 4(2):134–142, 2015.

[10] Youhao Li, Qujiang Lei, ChaoPeng Cheng, Gong Zhang, Weijun Wang, and Zheng Xu. A review: machine learning on robotic grasping. In *Eleventh International Conference on Machine Vision (ICMV 2018)*, volume 11041, page 110412U. International Society for Optics and Photonics, 2019.

[11] Tomas Lozano-Perez. Spatial planning: A configuration space approach. In *Autonomous robot vehicles*, pages 259–271. Springer, 1990.

[12] Thi Thoa Mac, Cosmin Copot, Duc Trung Tran, and Robin De Keyser. Heuristic approaches in robot path planning: A survey. *Robotics and Autonomous Systems*, 86:13–28, 2016.

[13] Ofir Marom and Benjamin Rosman. Belief reward shaping in reinforcement learning. In *Thirty-Second AAAI Conference on Artificial Intelligence*, 2018.

[14] Richard M Murray. *A mathematical introduction to robotic manipulation*. CRC press, 2017.

[15] Andrew Y Ng, Daishi Harada, and Stuart Russell. Policy invariance under reward transformations: Theory and application to reward shaping. In *ICML*, volume 99, pages 278–287, 1999.

[16] Franz Reuleaux. *The kinematics of machinery: outlines of a theory of machines*. Courier Corporation, 2013.

[17] John Greenlees Semple and Leonard Roth. *Introduction to algebraic geometry*. Oxford University Press, USA, 1985.

[18] S Tang, W Khaksar, N Ismail, and M Ariffin. A review on robot motion planning approaches. *Pertanika Journal of Science and Technology*, 20(1):15–29, 2012.

[19] Matej Večerík, Todd Hester, Jonathan Scholz, Fumin Wang, Olivier Pietquin, Bilal Piot, Nicolas Heess, Thomas Rothörl, Thomas Lampe, and Martin Riedmiller. Leveraging demonstrations for deep reinforcement learning on robotics problems with sparse rewards. *arXiv preprint arXiv:1707.08817*, 2017.

[20] Eric Wiewiora. Reward shaping. *Encyclopedia of Machine Learning*, pages 863–865, 2010.

[21] Eric Wiewiora, Garrison W Cottrell, and Charles Elkan. Principled methods for advising reinforcement learning agents. In *Proceedings of the 20th International Conference on Machine Learning (ICML-03)*, pages 792–799, 2003.

[22] Wikipedia. Taxicab geometry — Wikipedia, the free encyclopedia. http://en.wikipedia.org/w/index.php?title=Taxicab%20geometry&oldid=926081770, 2019. [Online; accessed 11-December-2019].

[23] Melonee Wise, Michael Ferguson, Derek King, Eric Diehr, and David Dymesich. Fetch and freight: Standard platforms for service robot applications. In *Workshop on autonomous mobile service robots*, 2016.

[24] Natsuki Yamanobe, Weiwei Wan, Ixchel G Ramirez-Alpizar, Damien Petit, Tokuo Tsuji, Shuichi Akizuki, Manabu Hashimoto, Kazuyuki Nagata, and Kensuke Harada. A brief review of affordance in robotic manipulation research. *Advanced Robotics*, 31(19-20):1086–1101, 2017.